\documentclass[10pt,twocolumn,letterpaper]{article}

\usepackage[pagenumbers]{cvpr} 
\usepackage{scalerel}  
\usepackage{times}
\usepackage{epsfig}
\usepackage{graphicx}
\usepackage{amsmath}
\usepackage{pifont}
\usepackage{amssymb}
\usepackage{booktabs}
\usepackage{times}
\usepackage{epsfig} 
\usepackage{makecell}
\usepackage{times}
\usepackage{silence}
\usepackage{epstopdf}
\usepackage{color, colortbl}
\usepackage{float}  
\usepackage{caption}
\usepackage{multirow} 
\usepackage{url}
\usepackage{cite}
\usepackage{placeins}
\usepackage[group-separator={,}]{siunitx}
\usepackage[utf8]{inputenc}
\usepackage{quoting,xparse}
\usepackage{paralist}
\usepackage{listings}
\usepackage{multicol} 

\NewDocumentCommand{\bywhom}{m}{
  {\nobreak\hfill\penalty50\hskip1em\null\nobreak
   \hfill\mbox{\normalfont(#1)}%
   \parfillskip=0pt \finalhyphendemerits=0 \par}%
}

\NewDocumentEnvironment{pquotation}{m}
  {\begin{quoting}[
     indentfirst=true,
     leftmargin=\parindent,
     rightmargin=\parindent]\itshape}
  {\bywhom{#1}\end{quoting}}

\newcommand{\dt}[1]{\fontsize{8pt}{0.1em}\selectfont (#1)}

\newcommand{\vspaceunderfig}{\vspace{-0.2cm}}

\newcommand{\comment}[1]{}

\definecolor{LightCyan}{rgb}{0.88,1,1}
\definecolor{Gray}{gray}{0.9}
\definecolor{Pink}{rgb}{1, 0, 1}
\definecolor{azure}{rgb}{0.0, 0.44, 1.0}
\definecolor{bleudefrance}{rgb}{0.19, 0.55, 0.91}
\definecolor{cobalt}{rgb}{0.0, 0.28, 0.67}
\definecolor{electricpurple}{rgb}{0.75, 0.0, 1.0}
\definecolor{cvprblue}{rgb}{0.21,0.49,0.74}
\definecolor{lightblue}{rgb}{0.85, 0.95, 1}
\definecolor{lightorange}{rgb}{1, 0.95, 0.85}
\definecolor{lightpink}{rgb}{1, 0.9, 0.95}
\usepackage[pagebackref,breaklinks,colorlinks,citecolor=teal]{hyperref}

\newcommand{\re}[1]{\textcolor{red}{#1}}

\newcommand{\bl}[1]{\textcolor{blue}{#1}}

\newcommand{\pp}[1]{\textcolor{purple}{#1}}

\newcommand{\pink}[1]{\textcolor{Pink}{#1}}
\definecolor{LightCyan}{rgb}{0.88,1,1}

\usepackage[capitalize]{cleveref}
\crefname{section}{Sec.}{Secs.}
\Crefname{section}{Section}{Sections}
\Crefname{table}{Table}{Tables}
\crefname{table}{Tab.}{Tabs.}
\usepackage[dvipsnames]{xcolor}
\usepackage{algorithm}
\usepackage{algorithmic}
\usepackage{color} 

\definecolor{codegreen}{rgb}{0,0.6,0}
\definecolor{codegray}{rgb}{0.5,0.5,0.5}
\definecolor{codepurple}{rgb}{0.58,0,0.82}
\definecolor{backcolour}{rgb}{0.95,0.95,0.92}

\lstdefinestyle{mystyle}{
    backgroundcolor=\color{backcolour},   
    commentstyle=\color{codegreen},
    keywordstyle=\color{magenta},
    numberstyle=\tiny\color{codegray},
    stringstyle=\color{codepurple},
    basicstyle=\ttfamily\footnotesize,
    breakatwhitespace=false,         
    breaklines=true,                 
    captionpos=b,                    
    keepspaces=true,                 
    numbers=left,                    
    numbersep=5pt,                  
    showspaces=false,                
    showstringspaces=false,
    showtabs=false,                  
    tabsize=2
}

\title{VCoder: Versatile Vision Encoders for Multimodal Large Language Models}

\author{Jitesh Jain\textsuperscript{1} \quad Jianwei Yang\textsuperscript{2} \quad Humphrey Shi\textsuperscript{1,3}\\
{\small \textsuperscript{1}SHI Labs @ Georgia Tech \quad \textsuperscript{2}Microsoft Research} \quad \small \textsuperscript{3}Picsart AI Research (PAIR)
\\
{\small \textbf{\url{https://github.com/SHI-Labs/VCoder}}
}
}

\begin{document}

\maketitle

\begin{abstract}

\noindent
Humans possess the remarkable skill of Visual Perception, the ability to see and understand the seen, helping them make sense of the visual world and, in turn, reason. Multimodal Large Language Models (MLLM) have recently achieved impressive performance on vision-language tasks ranging from visual question-answering and image captioning to visual reasoning and image generation. However, when prompted to identify or count (perceive) the entities in a given image, existing MLLM systems fail. Working towards developing an accurate MLLM system for perception and reasoning, we propose using Versatile vision enCoders (\textbf{VCoder}) as perception eyes for Multimodal LLMs. We feed the VCoder with perception modalities such as segmentation or depth maps, improving the MLLM's perception abilities. Secondly, we leverage the images from COCO and outputs from off-the-shelf vision perception models to create our COCO Segmentation Text (\textbf{COST}) dataset for training and evaluating MLLMs on the object perception task. Thirdly, we introduce metrics to assess the object perception abilities in MLLMs on our COST dataset. Lastly, we provide extensive experimental evidence proving the VCoder's improved object-level perception skills over existing Multimodal LLMs, including GPT-4V. We open-source our dataset, code, and models to promote research.
\end{abstract}    
\vspace{-0.3cm}
\section{Introduction}
\label{sec:intro}
\vspace{-0.3cm}

\noindent
\begin{pquotation}{GPT-4~\cite{openai2023gpt4}, 2023}
`Perception is the soil; reasoning, the seed. Without fertile ground, the seed cannot flourish.'
\end{pquotation}

\begin{figure}[t!]
\centering
\includegraphics[width=1\linewidth]{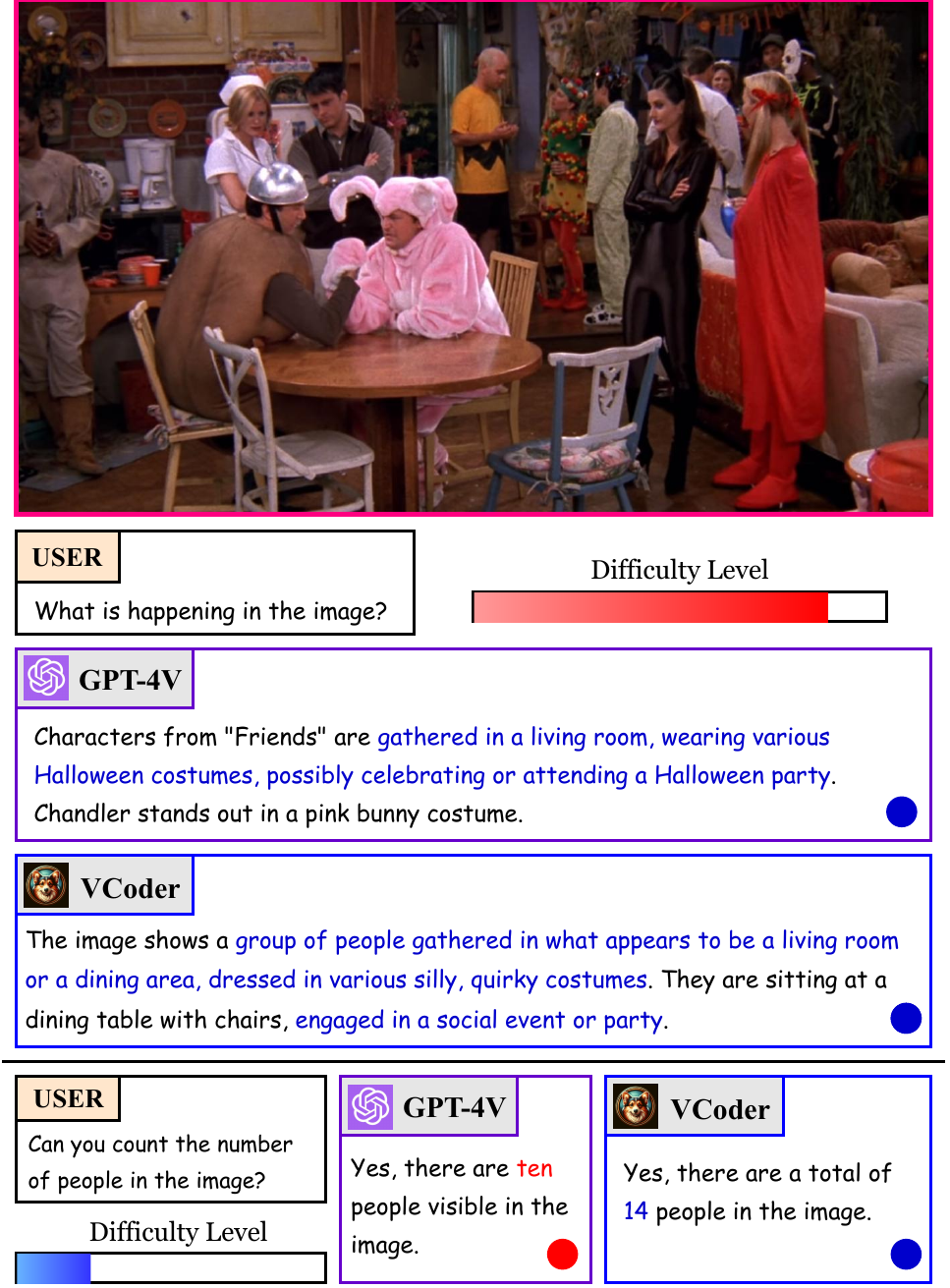} \\
\vspaceunderfig
\caption{GPT-4V~\cite{openai2023gpt4} (access date: Dec 16, 2023) returns impressive responses when prompted to describe complex visual scenes. However, it fails at the simple task of counting in the same scene. Our VCoder returns the correct count of people.}
\vspace{-0.5cm}
\label{fig:difficulty}
\end{figure}

\begin{figure*}[t!]
\centering
\includegraphics[width=\textwidth]{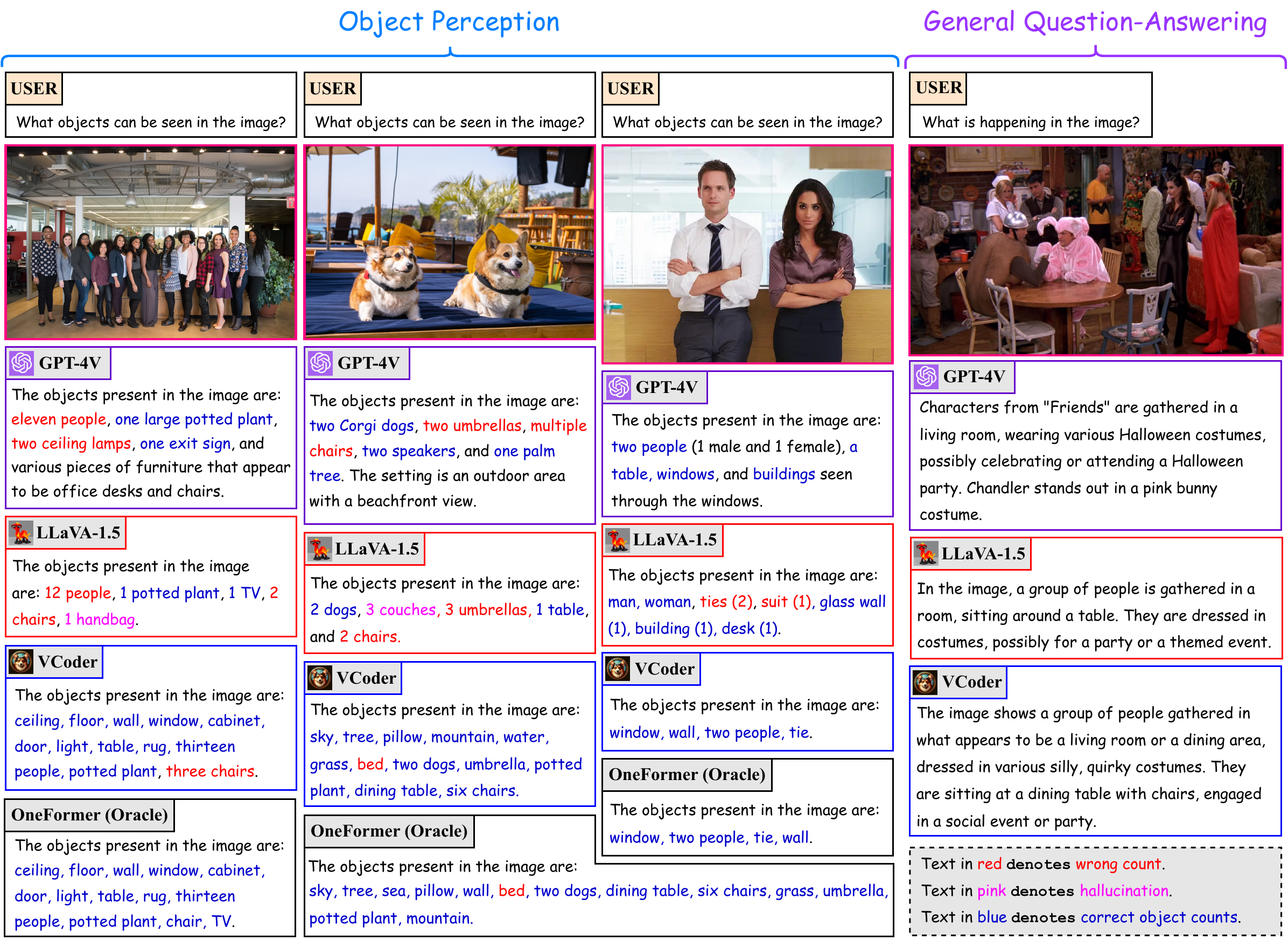}
\vspace{-0.6cm}
\captionof{figure}{\textbf{MLLMs counting and identifying objects.} As shown in the first column, GPT-4V~\cite{openai2023gpt4} (access date: Dec 16, 2023) and LLaVA-1.5~\cite{liu2023improvedllava} both fail at counting \textit{people}. Moreover, LLaVA-1.5~\cite{liu2023improvedllava} misses background entities like \textit{window, wall, etc.} and hallucinates about the presence of a handbag. VCoder can predict the \textit{people} counts and other background entities accurately except \textit{chairs}. Similarly, in the second column, GPT-4V and LLaVA-1.5 fail at counting \textit{chairs} while the VCoder matches the Oracle's performance. Notably, all MLLMs can perceive objects accurately for a non-cluttered image in the third column, with LLaVA-1.5 failing at counting ties. Our VCoder can also accurately perform general question-answering tasks, as shown in the fourth column. We treat OneFormer~\cite{jain2022oneformer} as the Oracle for object perception. 
\re{Red} text represents counting mistakes; \pink{pink} text represents hallucination; \bl{blue} text represents correct object perception.}
\vspaceunderfig
\label{fig:comparison}
\end{figure*}

\noindent
The ability to think and reason is one of the most remarkable traits that help humans function daily. Generally, understanding the environment precedes the act of thinking and reasoning~\cite{Kuhn1991SkillsOfArgument}. Following the success of ChatGPT-like instruction following AI agents~\cite{chatgpt, touvron2023llama2, vicuna2023, openai2023gpt4, falcon40b} at language understanding and reasoning, researchers have leveraged LLMs to develop instruct frameworks~\cite{zhu2023minigpt, liu2023llava, instructblip, ye2023mplugowl} that can understand vision and language inputs in an effort to imitate human perception and reasoning ability. We refer to such systems as Multimodal LLMs (MLLM). Although MLLMs exhibit the ability to perform complex vision-language tasks like visual captioning~\cite{liu2023improvedllava, alayrac2022flamingo, awadalla2023openflamingo}, image generation~\cite{Emu, koh2023generating, jin2023unified}, visual reasoning and grounding~\cite{zheng2023minigpt5, kosmos-2, kosmos-1}, they often display sub-par performance at simple tasks like counting objects (\cref{fig:difficulty}). As shown in \cref{fig:comparison}, MLLMs output incorrect object counts (\textit{people, chairs}) and hallucinate about the presence (\textit{handbag, couch}) of certain objects when prompted to identify entities in a visual input. The perception performance is much worse when the scenes are cluttered with many entities. Consequently, a natural question arises: \textit{``How to develop MLLM systems that respond to \textbf{perception} questions accurately?"}

This work aims to improve Multimodal LLMs at the simple yet fundamental object-level perception skills, including counting. Our motivation stems from the intuition that one can only describe and reason about a visual scene with the correct understanding of the entities in the image. In our effort to develop an accurate Multimodal LLM perception system, we face three significant challenges: (i) the scarcity of a vision-language dataset focused on the object perception task; (ii) existing open-sourced Multimodal LLMs usually use the ViT from CLIP~\cite{Radford2021LearningTV} with an RGB image as input as the visual component that majorly focuses only on salient objects, and (iii) the absence of evaluation metrics to quantitatively measure Multimodal LLMs' object perception and in particular, counting skills. We list our efforts to overcome the issues above in the following paragraphs.

The contemporary vision-language models~\cite{Radford2021LearningTV, li2023blip2, instructblip} owe their success to the availability of large-scale image-text datasets~\cite{sbu_captions, schuhmann2021laion400m, cc12m}. However, these datasets are more focused on image captioning~\cite{li2022blip} and VQA~\cite{agrawal2016vqa} tasks, making them unfit for training Multimodal LLMs for basic perception skills like object identification and counting. To overcome the scarcity of fundamental perception-focused image-text data, we leverage images from the COCO~\cite{coco} dataset and use predictions from off-the-shelf visual perception models~\cite{jain2022oneformer, oquab2023dinov2, DPT} to prepare a \textbf{CO}CO \textbf{S}egmentation \textbf{T}ext (\textbf{COST}) dataset comprising of question-answer pairs about the objects (background and foreground) present in each image. We provide more details in \cref{subsec:cost}.

Inspired by diffusion models that add various perception ``control" or ``context'' images~\cite{controlnet, xu2022versatile, xu2023prompt, mou2023t2i} as auxiliary inputs to aid image generation, we propose feeding extra perception modalities as control inputs through additional vision encoders, which we term as our Versatile vision enCoders (\textbf{VCoder}). In this work, we focus on the task of object perception and leverage a segmentation map, depth map, or both as the control inputs; however, the same design can be extended to other modalities. Our VCoder projects the control inputs' information into the LLM's space as shown in \cref{fig:vcoder}. We hypothesize that this added control helps the MLLM improve its object perception ability.

Lastly, owing to the absence of metrics to quantify the counting ability in MLLMs, we propose computing a count score (\textbf{CS}) using one-to-one matching of object words in the ground truth and MLLM's answer. We also compute a hallucination score (\textbf{HS}) based on the extra objects in the MLLM's response that are absent from the ground truth. Similarly, we introduce a depth score (\textbf{DS}) to quantify the object order prediction performance in MLLMs.

Among the open-source MLLMs, we choose LLaVA-1.5~\cite{liu2023improvedllava} as our base MLLM due to its impressive performance. Our extensive experimental analysis demonstrates the importance of our COST dataset and VCoder LLaVA-1.5's improved perception ability. To summarize, our contributions are as follows:

\begin{compactitem}
    \item We propose using extra (perception) control inputs and feeding those to a \textbf{V}ersatile en\textbf{Coder} (\textbf{VCoder}) for improved object perception performance.
    \item We introduce a COCO Segmentation Text (\textbf{COST}) dataset to train and evaluate Multimodal LLM systems on the fundamental object-level perception tasks of object identification, counting, and order prediction.
    \item Furthermore, to quantify the object perception ability in MLLMs, we propose calculating a count score (\textbf{CS}), a hallucination score (\textbf{HS}) and a depth score (\textbf{DS}). Our experiments show that the VCoder-adapted LLaVA-1.5 outperforms the baseline MLLMs on all metrics when validated on the COST dataset.
\end{compactitem}

\section{Related Work}
\label{sec:rel-work}

\subsection{Visual Perception}

The fundamental nature of visual perception makes it a critical component in MLLM systems. The task of perception can be divided into sub-tasks, including dense prediction tasks like image segmentation~\cite{image-parse, fcn, semask, jain2022oneformer} and depth estimation~\cite{eigen2014depth, DPT, girdhar2022omnivore}, and sparse prediction tasks like object detection~\cite{viola2001rapid,detr} and pose estimation~\cite{toshev2014deeppose, cootes2001active}. In the deep learning era, initial methods tackled the perception task using CNN based methods~\cite{lecun1998gradient,deeplabv1, mask-rcnn, panoptic-deeplab, faster-rcnn, laina2016deeper, toshev2014deeppose} with recent methods shifting to the use of vision transformer based architectures~\cite{jain2022oneformer, mask2former, segformer, DPT, girdhar2022omnivore, deformable-detr}. In this work, we tackle the fundamental task of object-level perception, mainly focusing on predicting names, counts, and order of objects in an image using MLLMs.

\subsection{Visual Understanding with LLMs}

Using LLMs for vision applications is not a new concept. In a nutshell, developing Multimodal LLMs involves projecting~\cite{li2023blip2, alayrac2022flamingo, awadalla2023openflamingo, frozen} the features from a vision encoder~\cite{clip, vit} to the embedding space of a language model (LLM)~\cite{vicuna2023, touvron2023llama, touvron2023llama2}, and, instruction-tuning on a vision-language dialog dataset.

LLaVA~\cite{liu2023llava} proposed a pipeline to convert existing image-text data into dialog format and then finetuned a CLIP~\cite{Radford2021LearningTV} and LLaMA~\cite{touvron2023llama} model end-to-end on their collected dataset showing one of the earliest evidence of visual-language instruction tuning. Concurrent to LLaVA, MiniGPT-4~\cite{zhu2023minigpt} used the visual encoder from BLIP2~\cite{li2023blip2} and used a linear layer for projecting visual features into Vicuna's~\cite{vicuna2023} feature space. InstructBLIP~\cite{instructblip} open-sourced a collection of 16 different datasets covering various vision tasks like VQA, reasoning, captioning, classification, etc., and finetuned a BLIP2 model on their dataset. mPLUG-Owl~\cite{ye2023mplugowl} proposed using a vision abstractor and finetuning the vision encoder. More recently, LLaVA-1.5~\cite{liu2023improvedllava} proposed using an MLP as the projector and finetuned on academic instruction datasets to achieve state-of-the-art performance on various benchmarks~\cite{hudson2018gqa, pope, fu2023mme, 2023opencompass}. Among various open-source MLLMs~\cite{qwen, kosmos-1, zeng2023matters, li2023otter, li2023mimicit}, we chose LLaVA-1.5 as our baseline due to its superior performance.

\begin{figure*}[t!]
\centering
\includegraphics[width=1\linewidth]{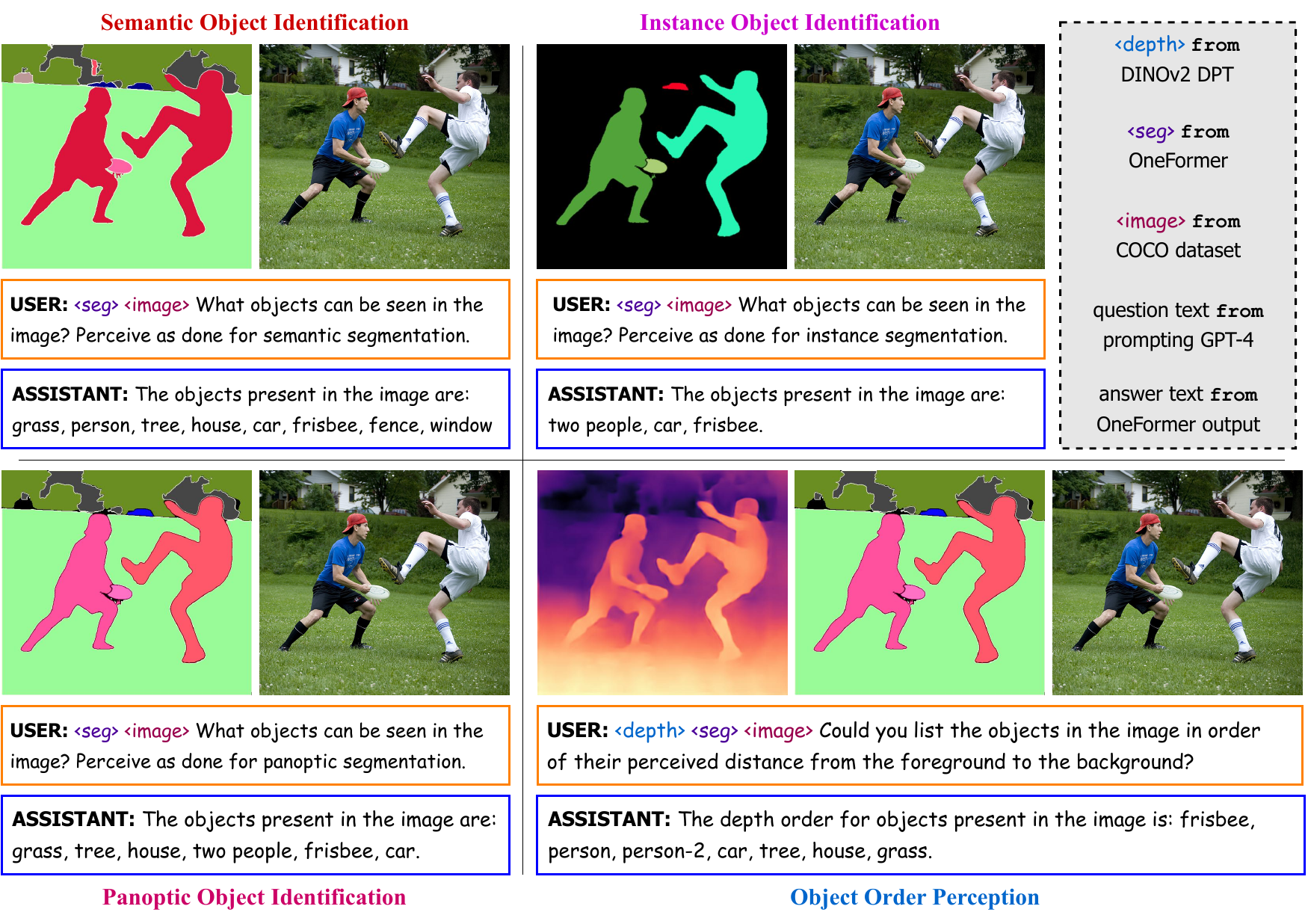} \\
\vspace{-0.3cm}
\caption{\textbf{Organization of the COST dataset}. We incorporate the images from COCO~\cite{coco}, the questions from GPT-4~\cite{openai2023gpt4}, and the segmentation outputs from OneFormer~\cite{jain2022oneformer} in a question-answer format for training and evaluating MLLMs on the object identification task. We also extend COST to the object order perception task by incorporating depth map outputs from DINOv2~\cite{oquab2023dinov2} DPT~\cite{DPT}. COST can be extended to more object-level tasks by similarly incorporating other modalities (for example, keypoint maps).}
\vspace{-0.4cm}
\label{fig:train_data}
\end{figure*}

\subsection{Perception Hallucination in MLLMs}

Since the introduction of LLMs, there has been a comprehensive study about their ability to hallucinate~\cite{zhang2023hallucination} in the NLP community. However, the phenomenon of hallucination in Multimodal LLMs has received comparatively less attention. LRV-Instruction~\cite{liu2023aligning} introduced a new instruction-tuning dataset containing 400k visual instructions to prevent hallucination in MLLMs and measured performance treating responses from GPT-4~\cite{openai2023gpt4} as ground truths. More recently, HallusionBench~\cite{liu2023hallusionbench} quantitatively benchmarked various failure modes in MLLMs that lead to hallucinations based primarily on logical consistency and reasoning. Unlike these works that tried to benchmark MLLMs mainly on VQA-type tasks, this paper focuses on the object-level hallucination in MLLMs. 

The two closest works to our objective are POPE~\cite{pope} and CHAIR~\cite{objectHallucination}. On the one hand, POPE~\cite{pope} tried to measure hallucination in MLLMs using a binary ``Yes"-``No" answer policy in response to questions based on the absence or presence of an object in the image. On the other hand, CHAIR~\cite{objectHallucination} focused on measuring hallucination in image captioning based on only words and not counts for the objects. In our work, we consider not only object words but also the corresponding count to compute an object-level count score and hallucination score.
\section{Object Identification with MLLMs}
\label{sec:method}

\begin{figure*}[t!]
\centering
\includegraphics[width=1\linewidth]{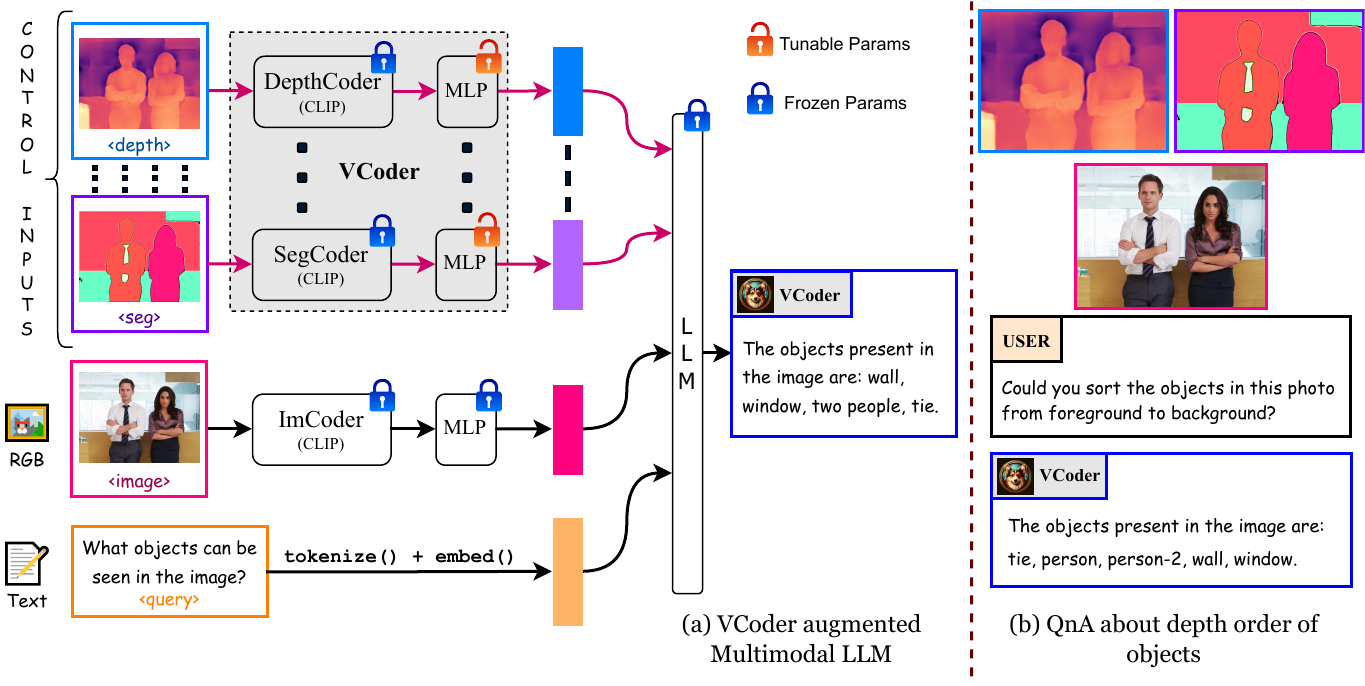} \\
\vspace{-0.3cm}
\caption{\textbf{Adapting Multimodal LLMs for accurate object perception with VCoder.} \textbf{(a)} We add our VCoder as an adapter to the LLaVA-1.5~\cite{liu2023improvedllava} and feed perception modalities as extra control inputs for improved object perception performance. During training, we freeze the components from LLaVA-1.5 (ImCoder, MLP, and LLM) to retain the original reasoning performance. \textbf{(b)} Using depth map and segmentation map as the control inputs to VCoder for the object order perception task.}
\label{fig:vcoder}
\vspace{-0.4cm}
\end{figure*}

Suppose you are invited to a Halloween party and want to bring candies for every person at that party. You ask your friend to send you a picture (\cref{fig:difficulty}) of the party room so that you can estimate the number of people and the number of candies you need to buy. In a hurry, you ask GPT-4V~\cite{openai2023gpt4}: ``\textit{Can you count the number of people in the image?}", and it responds: ``\textit{Yes, there are \re{ten} people visible in the image.}". Excited, you arrive at the party with ten candies but wait, you see fourteen people! Confused, you look at the image your friend sent you, and you can count \bl{fourteen} people in that image, realizing that GPT-4V fails at the simple task of counting the people in the picture. At the same time, it can accurately describe the happening of a Halloween party in the image (\cref{fig:difficulty}). We refer to the phenomenon of Multimodal LLMs failing at simple visual perception tasks while succeeding at complex visual reasoning tasks as Moravec's Paradox~\cite{moravec} in perception.

We hypothesize that one of the main reasons for the above phenomenon is the absence of conversations covering object identification for not only the salient objects but also the objects in the background from the instruction-tuning data for MLLMs. To overcome this issue, we prepare the COCO Segmentation Text (\textbf{COST}) dataset with COCO~\cite{coco} images and create sentences using the output from an image segmentation model~\cite{jain2022oneformer} to obtain an image-text dataset to train and evaluate MLLMs for object perception  MLLMs. Moreover, we also introduce a segmentation map as a control image input to the MLLM for better performance and quantify object perception performance with a count score (\textbf{CS}) and a hallucination score (\textbf{HS}).

\subsection{COST to Identify Objects with MLLMs}
\label{subsec:cost}

We find that image segmentation methods~\cite{jain2022oneformer, mask2former} can accurately identify salient (foreground objects like \textit{people, cars, etc.}) and background objects (like \textit{sky, wall, etc.}) in a given scene. Guided by this finding, we use images from the COCO~\cite{coco} dataset and obtain the corresponding segmentation outputs from OneFormer~\cite{jain2022oneformer}, a state-of-the-art image segmentation model. Next, we extract the object (class) names and counts from the segmentation outputs and convert them into a sentence form for the ground-truth answer:\textit{``The objects present in the image are: [$\text{CNT}_\text{1}$] [$\text{OBJ}_\text{1}$], [$\text{CNT}_\text{2}$] [$\text{OBJ}_\text{2}$], $\ldots$, [$\text{CNT}_\text{N}$] [$\text{OBJ}_\text{N}$]."}, with \textit{[$\text{OBJ}_i$]} representing the object name and \textit{[$CNT_i$]} representing the count (if greater than one) for the $i^{th}$ object in the image. We prompt GPT-4~\cite{openai2023gpt4} to collect a bucket of questions for three different object identification tasks: semantic, instance, and panoptic, corresponding to the three different image segmentation tasks. Finally, as shown in \cref{fig:train_data}, we organize the images from COCO, segmentation maps from OneFormer, questions from GPT-4, and sentences containing object information into a question-answer format to construct our \textbf{CO}CO \textbf{S}egmentation \textbf{T}ext (\textbf{COST}) dataset for training and evaluating MLLMs on the object identification task.

Statistically, we prompt GPT-4~\cite{openai2023gpt4} to return 20 questions for each question bucket (panoptic, semantic, and instance). In total, we used 280k images from the \texttt{train2017}, \texttt{test2017}, and \texttt{unlabeled2017} splits of the COCO~\cite{coco} dataset and corresponding segmentation outputs from OneFormer~\cite{jain2022oneformer} to form the visual component of the COST training dataset. Similarly, we prepare a COST validation split using the 5k images from the \texttt{val2017} split of the COCO dataset.

Note that a similar approach can extend the COST dataset to other perception modalities. In this work, we incorporate the depth map modality into our COST dataset for the object order perception task. Particularly, we leverage the publicly available DINOv2~\cite{oquab2023dinov2} DPT~\cite{DPT} model to obtain depth maps for COCO images and use the panoptic mask (from OneFormer~\cite{jain2022oneformer}) to estimate the depth order of objects in an image. We format the obtained ordering of objects into the text with the template: \textit{``The depth order for objects present in the image is: [$\text{OBJ}_\text{1}$], [$\text{OBJ}_\text{2}$], $\ldots$, [$\text{OBJ}_\text{J}$]."}, with \textit{[$\text{OBJ}_j$]} representing the $j^{th}$ object name. To maintain relative ordering among objects belonging to the same class, we append a count number to the second and later objects, as shown in the bottom right of \cref{fig:train_data} for \textit{person} and \textit{person-2}. Similar to the previous setting, we prompt GPT-4~\cite{openai2023gpt4} to return 20 questions for the object order perception task. We provide a detailed flow of obtaining ground-truth object orders in the appendix.

\subsection{VCoder for Multimodal LLMs}

We notice that existing open-source Multimodal LLMs generally use the ViT~\cite{vit} from CLIP~\cite{clip} as the image encoder (ImCoder) during instruction tuning. We reason that the ViT focuses mainly on salient objects because it is trained against captions, which leave out information about background regions. We argue that identifying objects in the background is critical for a Multimodal LLM to become skilled at perception. To overcome this limitation, we introduce a segmentation map as a control input~\cite{controlnet, mou2023t2i} into our Multimodal LLM. Specifically, we use the segmentation map from OneFormer~\cite{jain2022oneformer} and project it to the LLM's embedding space using a pretrained ViT~\cite{vit} (from CLIP~\cite{clip}) as a SegCoder and a two-layer MLP~\cite{liu2023improvedllava} which we collectively refer to as our \textbf{V}ersatile en\textbf{Coder} (\textbf{VCoder}). This extra control from the segmentation map results in considerable performance gains on the object identification task.

As shown in \cref{fig:vcoder}\textcolor{red}{a}, our VCoder adapted MLLM takes three sets of inputs: perception modalities as control inputs fed into the VCoder, an RGB image fed into an Image enCoder (and MLP), and the question from the user. The RGB image and text are tokenized to the \texttt{<img>} and \texttt{<query>} tokens, respectively. VCoder is flexible at handling various perception modalities with a special token for each modality. For example, the segmentation map and depth map inputs are tokenized to \texttt{<seg>} and \texttt{<depth>} tokens, respectively. Similarly, one can incorporate more modalities with modality-specific tokens. Finally, all tokenized embeddings are concatenated and fed into the LLM to obtain the final answer. We only use the \texttt{<seg>} input for the object identification task.

We treat our VCoder as an adapter, added to our base MLLM, LLaVA-1.5~\cite{liu2023improvedllava} to obtain the final MLLM framework for experiments. Note that we only train the MLP components in the VCoder on the COST dataset. We decided to keep all other parameters fixed during training to keep the reasoning ability unaffected while achieving improved object perception performance.

\subsection{Evaluating MLLMs for Object Identification}

Despite the availability of various metrics~\cite{pope, objectHallucination, lovenia2023nope} to measure object hallucination in vision-language models, no existing metric considers the explicit object counts while calculating their hallucination scores. We argue that object counts returned by an MLLM are a critical component that should not be overlooked while evaluating object identification performance. Therefore, we propose evaluating object identification performance in MLLMs using two metrics: count-score ($\mathbf{CS}$) and hallucination-score ($\mathbf{HS}$). 

\vspace{-0.5cm}
\begin{equation}
\begin{split}
G_\text{dict} & = \{\text{OBJ}^G_1: \text{CNT}^G_1; \cdots;  \text{OBJ}^G_N: \text{CNT}^G_N\} \\
P_\text{dict} & = \{\text{OBJ}^P_1: \text{CNT}^P_1; \cdots;  \text{OBJ}^P_M: \text{CNT}^P_M\} \\
\end{split}
\label{eq:dict_form}
\end{equation}
\vspaceunderfig

\begin{figure}[t!]
\centering
\includegraphics[width=1\linewidth]{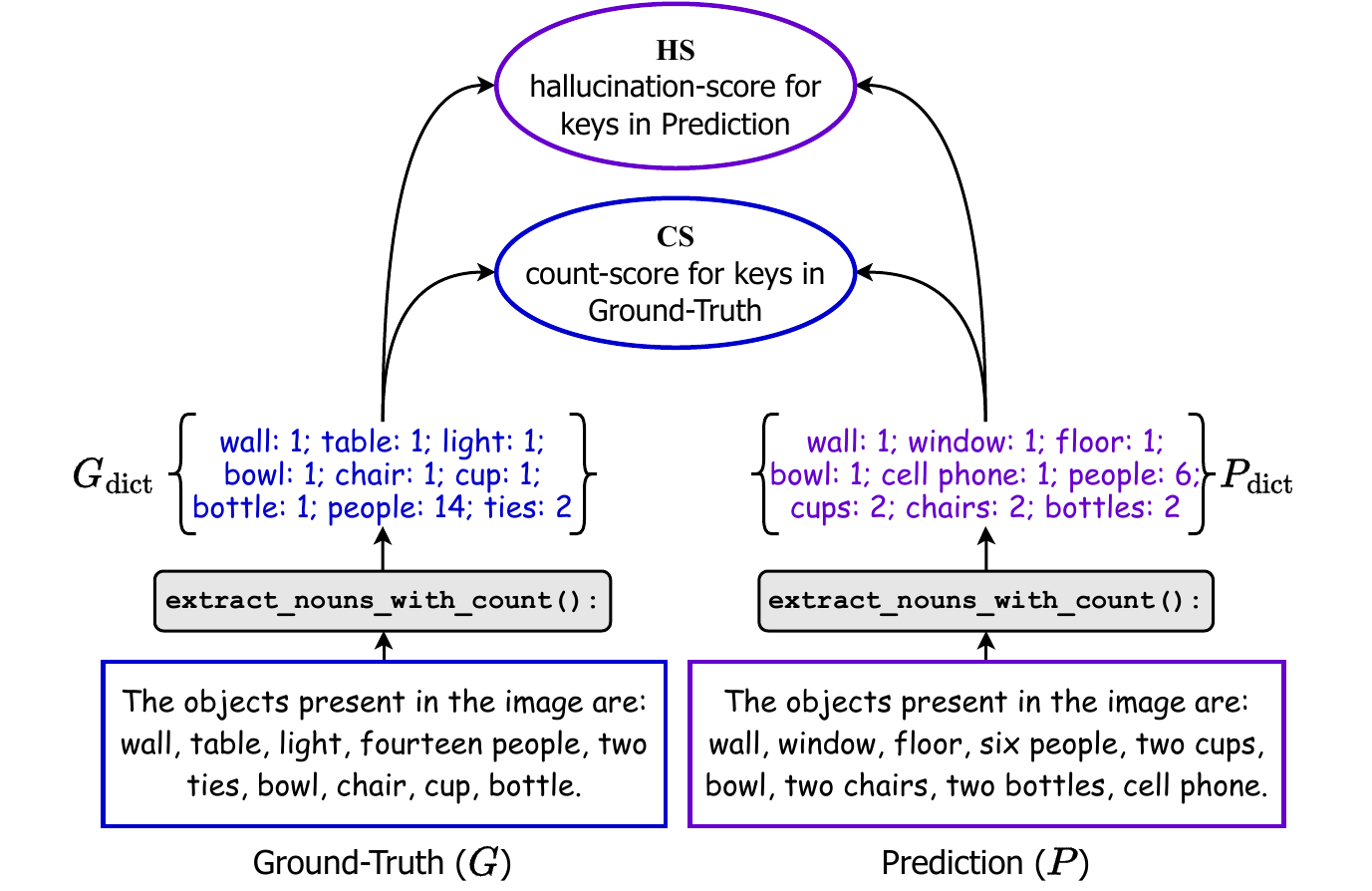} \\
\vspaceunderfig
\caption{\textbf{Evaluation Metrics for Object Identification}. We compare the object counts in the ground truth and prediction to calculate a count score (\textbf{CS}) and a hallucination score (\textbf{HS}).}
\vspace{-0.6cm}
\label{fig:eval}
\end{figure}

As shown in \cref{fig:eval}, given a ground-truth sentence ($G$) and an MLLM predicted response ($P$), we first extract the object words (nouns) and their corresponding count from both text samples and represent them in a dictionary form with keys as the object noun and the value as the corresponding object's count as shown in \cref{eq:dict_form} with $N$ and $M$ representing the number of different object nouns in the $G$ and $P$ respectively. Next, we perform one-to-one matching between the counts for keys with $G_\text{dict}$ and $P_\text{dict}$ as the reference for Count Score ($\mathbf{CS}$) and Hallucination Score ($\mathbf{HS}$), respectively, as shown in \cref{eq:score}.

\vspace{-0.5cm}
\begin{equation}
    \begin{split}
        & \mathbf{CS} = \frac{100}{N} \sum_{i=1}^{N} \begin{cases}
\frac{\min(\text{CNT}^G_i, \text{CNT}^P_i)}{\max(\text{CNT}^G_i, \text{CNT}^P_i)} & \text{if } I(\text{OBJ}^G_i, P_\text{dict}) \\
0 & \text{otherwise} \end{cases} \\
    & \mathbf{HS} = \frac{100}{M} \sum_{j=1}^{M} \begin{cases}
1 - \frac{\min(\text{CNT}^P_j, \text{CNT}^G_j)}{\max(\text{CNT}^P_j, \text{CNT}^G_j)} & \text{if } I(\text{OBJ}^P_j, G_\text{dict}) \\
1 & \text{otherwise} \end{cases} \\
    & I(\text{OBJ}, D) = \begin{cases}
\text{True} & \text{if } \text{OBJ} \text{ is in } \texttt{keys}(D) \\
\text{False} & \text{otherwise} \end{cases} \\
    \end{split}
\label{eq:score}
\end{equation}
\noindent
\textbf{Count Score (CS)}. It represents the percentage of correct object counts predicted by the MLLM with respect to the ground-truth sentence. The higher the \textbf{CS}, the better.

\noindent
\textbf{Hallucination Score (HS)}. It represents the percentage of extra object counts predicted by the MLLM that do not exist in the ground-truth sentence. The lower the \textbf{HS}, the better.

Note that due to the one-to-one word-matching nature of our evaluation, we manually define a mapping between the categories in COCO~\cite{coco} and their synonyms~\cite{objectHallucination, Lu2018Neural}. For example, we replace words like \textit{man, woman, child, kid, boy, girl}, \emph{etc.} with the word \textit{person} in the MLLM's response before evaluation.

\begin{table*}[t!]
\centering
\scalebox{1.0}{
\begin{tabular}{l| c| cc| cc| cc}
 & {} & \multicolumn{2}{c|}{Semantic} & \multicolumn{2}{c|}{Instance} & \multicolumn{2}{c}{Panoptic} \\

 {Method} & Input Tokens & CS ($\uparrow$) & HS ($\downarrow$) & CS ($\uparrow$) & HS ($\downarrow$) & CS ($\uparrow$) & HS ($\downarrow$) \\

\midrule
\multicolumn{8}{l}{\textit{Closed Model, Open API}}  \\
\midrule

GPT-4V~\cite{openai2023gpt4} & \textit{$\langle$img$\rangle$} + \textit{$\langle$query$\rangle$}  & --- & --- & --- & --- & 38.4 & 83.0  \\

\midrule
\multicolumn{8}{l}{\textit{Existing Open-Source Multimodal LLMs}}  \\
\midrule

MiniGPT-4 LLaMA-2-7b~\cite{zhu2023minigpt} & \textit{$\langle$img$\rangle$} + \textit{$\langle$query$\rangle$}  & 6.2 & 92.2 & 5.6 & 97.7 & 6.2 & 94.9  \\

InstructBLIP Vicuna-7b~\cite{instructblip} & \textit{$\langle$img$\rangle$} + \textit{$\langle$query$\rangle$}  & 14.2 & 85.8 & 25.3 & 91.9 & 17.5 & 91.2  \\

LLaVA-1.5-7b~\cite{liu2023improvedllava} & \textit{$\langle$img$\rangle$} + \textit{$\langle$query$\rangle$}  & 30.6 & 60.1 & 50.3 & 75.9 & 38.7 & 67.3  \\

LLaVA-1.5-13b~\cite{liu2023improvedllava} & \textit{$\langle$img$\rangle$} + \textit{$\langle$query$\rangle$}  & 25.0 & 69.3 & 49.9 & 75.0 & 35.8 & 68.6  \\

CogVLM-17b~\cite{wang2023cogvlm} & \textit{$\langle$img$\rangle$} + \textit{$\langle$query$\rangle$}  & 33.4 & 67.5 & 43.5 & 86.2 & 40.6 & 75.9  \\

\midrule
\multicolumn{8}{l}{\textit{Baselines trained on the \textbf{COST} dataset}}  \\
\midrule
COST IT LLaVA-1.5-7b & \textit{$\langle$img$\rangle$} + \textit{$\langle$query$\rangle$} & 78.7 & 22.1 & 67.5 & 30.3 & 71.9 & 28.2 \\

Soft-Prompted LLaVA-1.5-7b & \textit{$\langle$prompt$\rangle$} + \textit{$\langle$img$\rangle$} + \textit{$\langle$query$\rangle$} & 36.2 & 56.7 & 18.4 & 72.2 & 26.8 & 63.0  \\

ImCoder LLaVA-1.5-7b  & \textit{$\langle$img$\rangle$} + \textit{$\langle$img$\rangle$} + \textit{$\langle$query$\rangle$} & 78.9 & 22.7 & 64.0 & 29.4 & 70.8 & 27.9  \\

\midrule
\multicolumn{8}{l}{\textit{\textbf{VCoder} augmented LLaVA-1.5}}  \\
\midrule

\textbf{VCoder} LLaVA-1.5-7b &  \textit{$\langle$seg$\rangle$} + \textit{$\langle$img$\rangle$} + \textit{$\langle$query$\rangle$} & \underline{88.6} & \underline{10.4} & \underline{71.1} & \underline{26.9} & \underline{86.0} & \underline{12.8}  \\
\textbf{VCoder} LLaVA-1.5-13b &  \textit{$\langle$seg$\rangle$} + \textit{$\langle$img$\rangle$} + \textit{$\langle$query$\rangle$} & \textbf{89.0} & \textbf{10.0} & \textbf{73.3} & \textbf{25.0} & \textbf{87.2} & \textbf{11.6}  \\

\bottomrule
\end{tabular}}
\vspaceunderfig
\caption{\textbf{Comparison to baseline Multimodal LLMs on the COST validation dataset for Object Identification.} We compare our VCoder to existing off-the-shelf baseline MLLMs: MiniGPT-4~\cite{zhu2023minigpt}, InstructBLIP~\cite{instructblip}, LLaVA-1.5~\cite{liu2023improvedllava}, and CogVLM~\cite{wang2023cogvlm}. We also train three different variants of LLaVA-1.5 on the COST dataset: \textit{COST IT} mixes the COST training data with the instruction tuning data; \textit{Soft-Prompted} uses a set of learnable tokens, and \textit{ImCoder} uses an RGB image as the control input. Our \textbf{VCoder} adapted LLaVA-1.5 performs the best on all three object perception tasks. Note: \textit{$\langle\cdot\rangle$} denotes input tokens to LLM with \textit{seg} representing segmentation map, \textit{img} representing RGB image, \textit{prompt} representing learnable prompt, and \textit{query} representing the user question. We also evaluate the performance of GPT-4V~\cite{openai2023gpt4} on the COST dataset using the publicly accessible paid API released by OpenAI. Our VCoder-adapted LLaVA-1.5 shows the best performance on object identification among all MLLMs.}
\vspaceunderfig
\label{tab:eval_scores}
\end{table*}
\section{Experiments}
\label{sec:exp}

We use LLaVA-1.5~\cite{liu2023improvedllava} as our base MLLM. LLaVA-1.5 uses CLIP-ViT-L-336px~\cite{clip} as the image encoder (ImCoder) with a two-layer MLP as projection and Vicuna-1.5~\cite{zheng2023judging} as the LLM. Inside our VCoder, we also use a CLIP-ViT-L-336px to encode the control inputs and project the features into the LLM embedding space using modality-specific two-layer MLPs. We resize the visual inputs to 336$\times$336 resolution (corresponds to 576 tokens) for our MLLM. During training, we load the instruction-tuned weights from LLaVA-1.5 and keep those frozen while only tuning the MLP component of our VCoder. We use the publicly available OneFormer~\cite{jain2022oneformer} model trained on COCO~\cite{coco} with DiNAT-L~\cite{dinat,hassani2023neighborhood} backbone to obtain the segmentation map. For getting depth maps, we use the publicly available ViT-L/14 distilled variant of DINOv2~\cite{oquab2023dinov2} DPT~\cite{DPT} trained on the NYUd~\cite{nyud} dataset. In this section, we discuss our results on the object identification task. Please refer to \cref{sec:order} for our results on the object order perception task. 

\subsection{Implementation Details}

\vspace{0.1cm}
\noindent
\textbf{Training Details.} We train our VCoder-adapted LLaVA-1.5 framework for two epochs on the COST training dataset with a batch size 256 and a learning rate of 1$e^{-3}$. For other training hyperparameters, we follow the settings used during the instruction-tuning stage in LLaVA-1.5~\cite{liu2023improvedllava}. Following \cite{jain2022oneformer}, we uniformly sample each object identification task (semantic, instance, and panoptic) during training. We also use the corresponding segmentation map from OneFormer~\cite{jain2022oneformer} as input to the VCoder during training and inference. On 8 A100 GPUs, it takes 8 and 14 hours to train our VCoder with the 7b and 13b variants of LLaVA-1.5 as the base MLLM, respectively. 

\vspace{0.1cm}
\noindent
\textbf{Evaluation Details.} We evaluate all MLLMs on the COST validation set. We separately evaluate semantic, instance, and panoptic object identification tasks while randomly sampling questions from the corresponding task's question bucket. Note that for evaluating all off-the-shelf MLLMs, we experiment with various prompts and finally use the prompt: \textit{``[QUESTION]. Return the answer in the paragraph format: `The objects present in the image are: ...' and then list the objects with their count in word format (if greater than 1) in front of them, like 'two people'."}, where [QUESTION] is the randomly sampled question from the object identification task bucket.

\subsection{Main Results}

\vspace{-0.1cm}
\textbf{Baselines.} We compare the performance of VCoder to open-source Multimodal LLMs, namely, MiniGPT-4~\cite{zhu2023minigpt}, InstructBLIP~\cite{instructblip}, LLaVA-1.5~\cite{liu2023improvedllava}, and CogVLM~\cite{wang2023cogvlm} on the COST validation set in \cref{tab:eval_scores}. Furthermore, we also provide three additional baselines, all trained for two epochs:

\noindent
\underline{COST IT LLaVA-1.5}: We mix the COST training data with the instruction tuning data used in LLaVA-1.5~\cite{liu2023improvedllava} and finetune a LLaVA-1.5 model from scratch following the settings from  Liu \emph{et al.}~\cite{liu2023improvedllava}.

\noindent
\underline{Soft-Prompted LLaVA-1.5}: We prepend 576 learnable tokens (\textit{$\langle$prompt$\rangle$}) to the LLM input and tune only the \textit{$\langle$prompt$\rangle$} parameters on the COST training dataset.

\noindent
\underline{ImCoder LLaVA-1.5}: We use an RGB image as the control input instead of a segmentation map and train VCoder on the COST training dataset.

As shown in \cref{tab:eval_scores}, we notice that all existing MLLMs perform poorly on our COST validation set, demonstrating their inability to count and identify objects accurately. Note that existing MLLMs perform relatively better on instance object identification, reaffirming our claim that MLLMs are better at detecting salient objects than background objects. Although the baselines trained on the COST dataset perform relatively better, they still lag in performance compared to the VCoder. Notably, a segmentation map performs considerably better than using an RGB image as the control input, proving the segmentation map's vitality.

\noindent
\textbf{Comparison to GPT-4V~\cite{openai2023gpt4}.} We utilize OpenAI's newly released \texttt{gpt-4-vision-preview}\footnote{\href{https://platform.openai.com/docs/guides/vision}{\texttt{https://platform.openai.com/docs/guides/vision}}} API to obtain responses from GPT-4V. Our experiments show that GPT-4V's responses are consistent across all object identification tasks, closely aligning with the panoptic identification task. Therefore, we compare our VCoder to GPT-4V only on the panoptic object identification to reduce API requests due to a daily limit of 500 API requests during this project. As shown in \cref{tab:eval_scores}, GPT-4V~\cite{openai2023gpt4} lags behind our VCoder by a considerable margin, reaffirming our claim that existing MLLMs cannot perform accurate object-level perception.
\begin{table}[t!]
\centering
\scalebox{1.0}{
\begin{tabular}{l| c}
 {Method} & Depth Score ($\downarrow$) \\
 
\midrule

LLaVA-1.5-7b~\cite{liu2023improvedllava}  & 166.1  \\

LLaVA-1.5-13b~\cite{liu2023improvedllava} & 227.2 \\

\midrule

\textbf{VCoder}-DS LLaVA-1.5-7b & \textbf{65.9}  \\

\textbf{VCoder}-DS LLaVA-1.5-13b & \textbf{63.3}  \\

\bottomrule
\end{tabular}
}
\vspaceunderfig
\captionof{table}{\textbf{Performance on Object Order Perception.} Our VCoder LLaVA-1.5 considerably outperforms LLaVA-1.5~\cite{liu2023improvedllava}, owing to the usage of control inputs and training on the COST dataset.}
\label{tab:depth_score}  
  \vspace{-0.5cm}
\end{table}

\section{Object Order Perception with MLLMs}
\label{sec:order}

As shown in \cref{fig:vcoder}, multiple perception modalities can be leveraged to improve object perception in MLLMs with our VCoder. This section presents our experiments with our VCoder using the segmentation and depth maps as the control inputs. We term the resulting MLLM as VCoder-DS LLaVA-1.5. Intuitively, predicting the object order implicitly means identifying the objects in an image. Therefore, for the object order perception task (\cref{fig:vcoder}\textcolor{red}{b}), we use both \texttt{<depth>} and \texttt{<seg>} inputs, while only the \texttt{<seg>} input as the additional control for object identification.

During training, we use a mixture of datasets, including the object identification and object order perception components from the COST dataset. We also use about 200k image-conversation (along with the corresponding segmentation map obtained using OneFormer~\cite{jain2022oneformer}) pairs randomly sampled from the instruction tuning data used in LLaVA-1.5~\cite{liu2023improvedllava}. We train our VCoder for one epoch following the same hyperparameter settings mentioned in \cref{sec:exp}.

As shown in \cref{tab:depth_score}, our VCoder-DS LLaVA-1.5 significantly outperforms the base MLLM, LLaVA-1.5~\cite{liu2023improvedllava} on the COST validation set. For quantitatively evaluating the performance of MLLMs on the depth order perception task, we calculate a depth score (\textbf{DS}) using the absolute difference between the position of objects in the ground truth and prediction. We provide more details about computing the depth score in the appendix.

\begin{figure}
\centering
\includegraphics[width=1.0\linewidth]{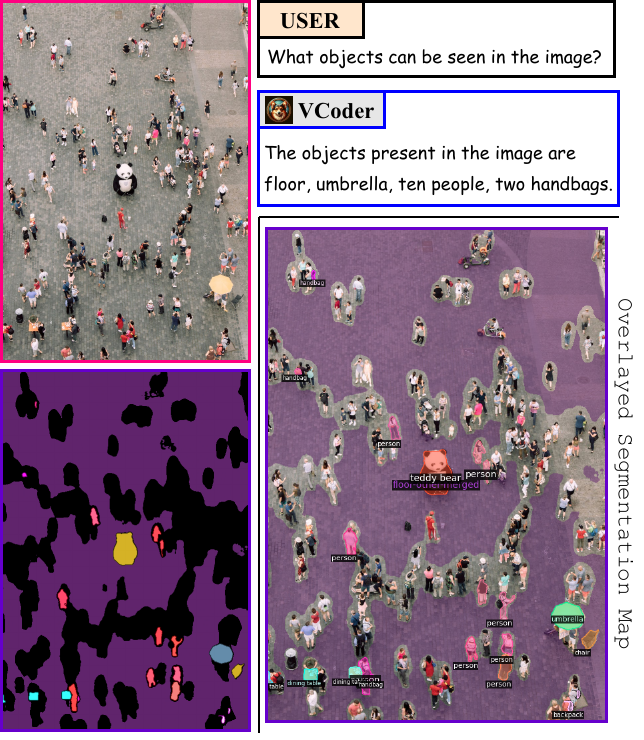}
\vspace{-0.7cm}
\caption{\textbf{Failure Case.} VCoder returns the wrong response when the input segmentation mask (control input) is inaccurate.}
\label{fig:fail}  
\vspace{-0.5cm}
\end{figure}

\section{Limitations}

Despite the improved object perception performance after training our VCoder on the COST dataset, certain limitations remain to be addressed for future work. Firstly, we build our COST dataset using OneFormer~\cite{jain2022oneformer}, which can only perceive objects belonging to a limited number of categories due to being trained on a closed-set vocabulary dataset~\cite{coco}. For real-world applications, it is imperative to develop an object perception benchmark for MLLMs covering many more classes with varying granularity than those in the COCO~\cite{coco}. Secondly, the count, hallucination, and depth scores use one-to-one word matching, which requires manually defining a mapping between synonymous words. It will be promising to explore ways to overcome manually defined synonym mappings. Lastly, as shown in \cref{fig:fail}, the inaccuracy in the segmentation map may result in the VCoder's failure. Exploring ways to reduce the over-dependency on control inputs to handle inaccurate context from the perception modalities would be interesting.

\section{Conclusion}

This work analyzes the object-level perception skills of Multimodal Large Language Models (VLMMs). Although MLLMs are good visual reasoners, they need to improve at the simple yet fundamental task of object perception. To improve object perception ability in MLLMs, we propose the COST dataset for training and evaluating MLLMs at the object perception task. We benchmark different off-the-shelf MLLMs and GPT-4V on our COST dataset and observe their lousy performance. Consequently, we propose using perception modalities as control inputs and a Versatile vision enCoders (\textbf{VCoder}) as an adapter for projecting the control inputs to the LLM embedding space. Our VCoder can easily be extended to leverage various modalities as the control inputs depending on the task. To quantify the object-level perception ability in MLLMs, we introduce a Count-Score (\textbf{CS}), a Hallucination-Score (\textbf{HS}), and a Depth-Score (\textbf{DS}). We adapted LLaVA-1.5 with VCoder, only trained the VCoder on our COST dataset, and demonstrated its improved performance at the object perception task while retaining the reasoning performance. We hope our work can inspire the research community to focus on developing object perception datasets for MLLMs and develop vision systems that are equally good at perception and reasoning in the future.

\paragraph{Acknowledgements.} We would like to extend our gratitude to Eric Zhang and Kai Wang (JJ's labmates) for an insightful discussion before the start of the project and valuable feedback on the design of Figure 2. We also thank the ML Center at Georgia Tech for generously supporting this work.
{
    \small
    \bibliographystyle{ieeenat_fullname}
    \bibliography{main}
}

\clearpage
\setcounter{page}{1}

\newpage
\appendix
\begin{center}{\bf \Large Appendix}\end{center}\vspace{-2mm}
\renewcommand{\thetable}{\Roman{table}}
\renewcommand{\thefigure}{\Roman{figure}}
\setcounter{table}{0}
\setcounter{figure}{0}

\Crefname{appendix}{Appendix}{Appendixes}

In this appendix, we first present our analysis of the effect of the quality of the segmentation map (control input) on the VCoder's performance in \cref{sec:mask_ablat}. Next, we provide details about obtaining ground-truth texts for the object order perception task along with the process to compute the depth score in \cref{sec:obj_order}. Lastly, we share analysis on the per-image object counts about the COST dataset in \cref{sec:cost_dataset}.  

\section{Control Through Segmentation Map}
\label{sec:mask_ablat}

We study the effect of segmentation map quality on object identification performance. Specifically, instead of using DiNAT-L OneFormer~\cite{jain2022oneformer} to obtain the segmentation map, we use the relatively worse segmentation models: ResNet-50~\cite{resnet} based Mask R-CNN~\cite{mask-rcnn}, Panoptic-FPN~\cite{pan-fpn}, and Swin-L~\cite{swin-T} based Mask2Former~\cite{mask2former} for the instance and panoptic object identification task, respectively. As shown in \cref{tab:ablat_mask}, we notice a considerable drop in performance with maps from Mask R-CNN and Panoptic FPN. However, the drop in performance is much lower with maps from a relatively newer and better Mask2Former model, demonstrating the importance of the segmentation map's quality.
\section{Object Order Perception}
\label{sec:obj_order}

In this section, we present the process of obtaining the ground truth ordering of objects in an image using segmentation and depth maps. Then, we share details about the logic used to compute the depth score (\textbf{DS}).

\subsection{Obtaining Ground Truth}

\begin{figure}[t!]
\centering
\includegraphics[width=1\linewidth]{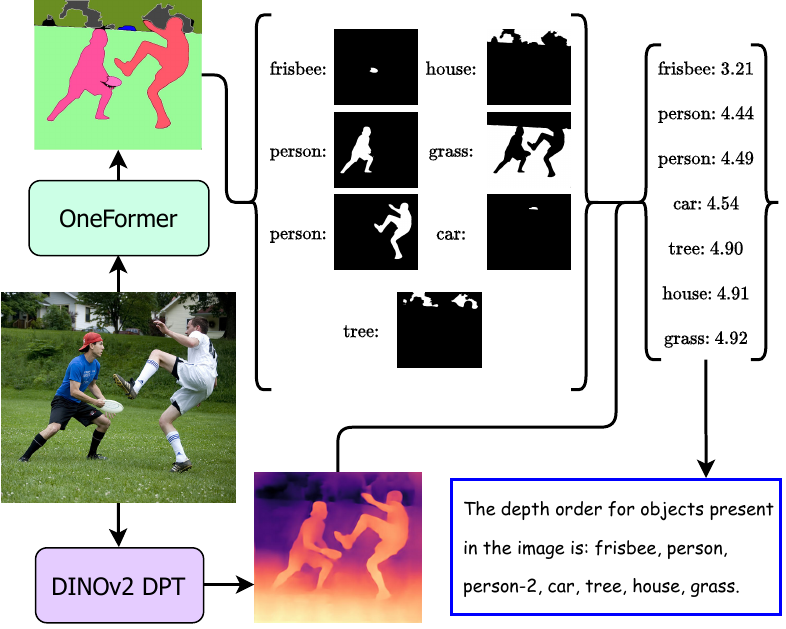} \\
\vspaceunderfig
\caption{\textbf{Data Engine to obtain Object Order GT}. We calculate the maximum pixel value inside each object's region using the depth and segmentation maps. We sort the obtained values in an ascending order to obtain the final object order.}
\vspace{-0.4cm}
\label{fig:depth_data_engine}
\end{figure}

To obtain the ground truth order for objects in an image, we utilize the fact that each pixel in a depth map (from DINOv2~\cite{oquab2023dinov2} DPT~\cite{DPT}) represents the distance~\cite{nyud} of that pixel from the camera. Therefore, as shown in \cref{fig:depth_data_engine}, we use the binary object masks (from OneFormer's~\cite{jain2022oneformer} panoptic prediction) to first obtain the corresponding regions in the depth map. Next, for each object region, we calculate the maximum pixel value representing the distance of the object's farthest point from the camera. Finally, we sort the values obtained in the previous in an ascending order to obtain the final order, starting with the closest object and ending with the farthest object. As mentioned in \cref{sec:order}, we append a number to the object name to represent the relative order of objects belonging to the same category.   

\subsection{Depth Score}
 In \cref{fig:depth_score_code}, we share the \texttt{python} code to compute the depth score given the ground truth and prediction for object orders in an image. Particularly, we first obtain the position of objects belonging to all categories and then compute the absolute difference using the position values for objects belonging to the same category in the ground truth and prediction. Note that to handle different numbers of objects in the prediction and ground truth, we use the position value as 100 for unmatched objects. We average the obtained score over all images to obtain the final depth score.

 \begin{table}[t!]
\centering
\scalebox{0.9}{
\begin{tabular}{l|c| cc}

 {Seg Model} & Year & CS ($\uparrow$) & HS ($\downarrow$) \\
\midrule
\multicolumn{4}{l}{\textit{Instance Object Identification}}  \\
\midrule

OneFormer~\cite{jain2022oneformer} & CVPR 2023 & \textbf{71.1} & \textbf{26.9}  \\
Mask R-CNN~\cite{mask-rcnn} & ICCV 2017 & 61.9 \pp{\dt{-9.2}} & 39.8 \pp{\dt{+12.9}} \\

\midrule
\multicolumn{3}{l}{\textit{Panoptic Object Identification}} \\
\midrule

OneFormer~\cite{jain2022oneformer} & CVPR 2023 & \textbf{86.0} & \textbf{12.8} \\
Mask2Former~\cite{mask2former} & CVPR 2022 & {76.5} \pp{\dt{-9.5}} & {26.1} \pp{\dt{+13.3}} \\
Panoptic FPN~\cite{pan-fpn} & CVPR 2019 & {64.2} \pp{\dt{-21.8}} & {33.3} \pp{\dt{+20.5}} \\

\bottomrule
\end{tabular}}
\vspaceunderfig
\caption{\textbf{Ablation on Quality of Segmentation Map}. Using segmentation maps from older models like Mask R-CNN~\cite{mask-rcnn} and Panoptic-FPN~\cite{pan-fpn} as the control input results in a performance drop due to the relatively low quality of the maps.}
\label{tab:ablat_mask}
\end{table}

\begin{figure*}
\begin{lstlisting}[language=Python]
def calculate_per_image_depth_score(gt, pred):
    position_gt, order_num = _get_order(gt)
    position_pred, _ = _get_order(pred)
    depth_distance = []
    
    for object in position_gt.keys():
        if position_pred is not None and object in position_pred.keys():
            order_pred = position_pred[object]
            order_gt = position_gt[object]
            # pad the object specific position list to make with 100 to make them equal for prediction and ground-truth
            if len(order_gt) < len(order_pred):
                order_gt.extend([100] * (len(order_pred) - len(order_gt)))
            elif len(order_pred) < len(order_gt):
                order_pred.extend([100] * (len(order_gt) - len(order_pred)))
            for i, j in zip(order_gt, order_pred):
                depth_distance.append(abs(i - j))
        else:
            depth_distance.append(100)  
    # normalize the score based on the total number of objects in the image
    return sum(depth_distance) / order_num

# helper function to calculate the order position of the objects in the image
def _get_order(text):
    order_num = 1  # order number of the object
    positions = {}
    # obtain object nouns
    nouns = _obtain_nouns(text)
    for noun in nouns:
        # obtain only object noun (person) from words like person-2
        object = noun.split("-")[0].strip()
        if object not in positions.keys():
            positions[object] = [order_num]
        else:
            positions[object].append(order_num)
        order_num += 1 
    return positions, order_num - 1
\end{lstlisting}
\vspace{-0.4cm}
\caption{Computing Depth Score for a given Image.}
\label{fig:depth_score_code}
\end{figure*}
\section{Object Counts in COST Dataset}
\label{sec:cost_dataset}

\begin{figure*}
\centering
\includegraphics[width=0.8\linewidth]{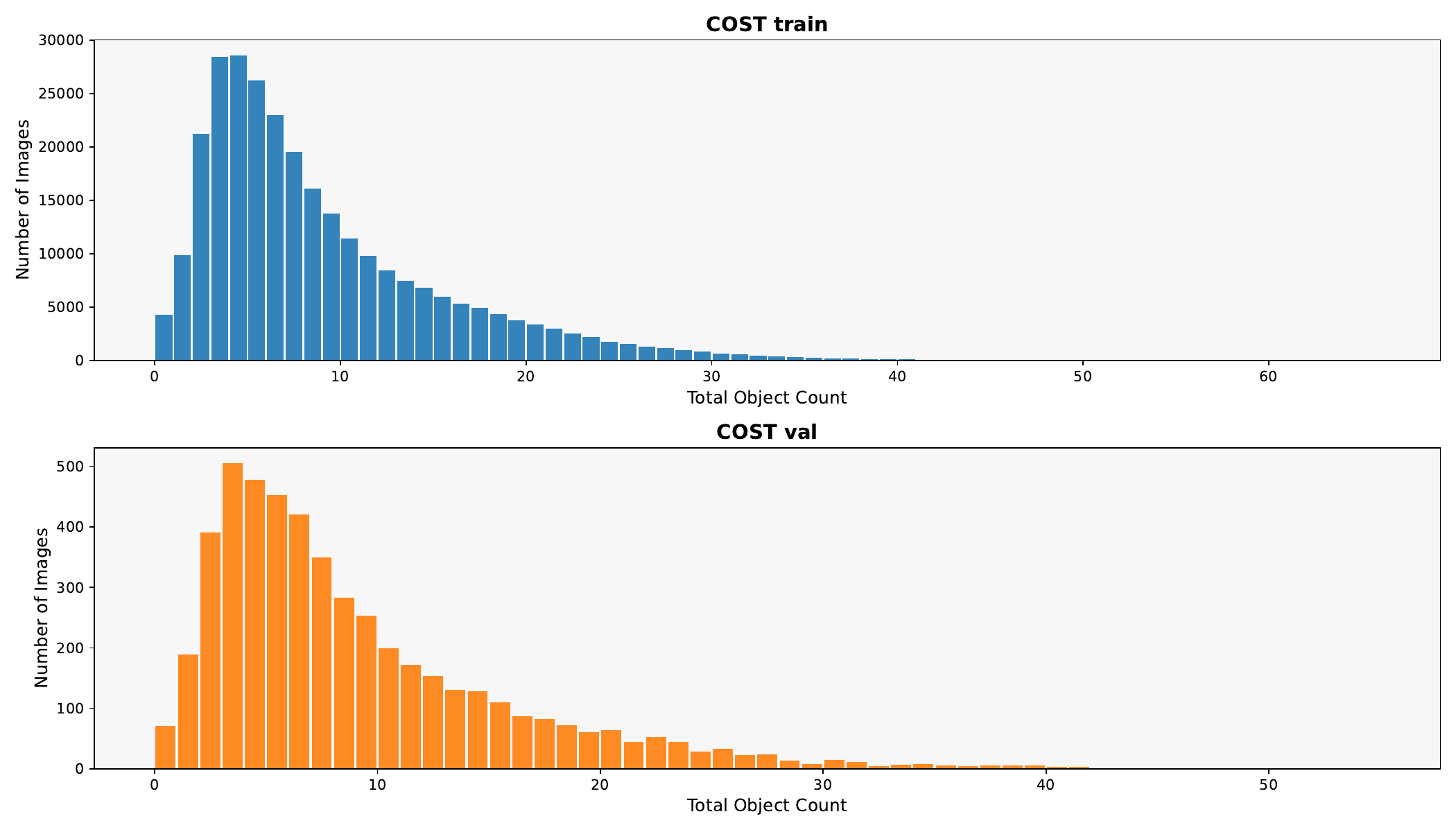} \\
\vspace{-0.3cm}
\caption{\textbf{Total Object Counts per image in the COST \texttt{train} and \texttt{val} splits.} We observe that our COST dataset does not include images with more than 60 objects and has a long tail beyond the object count of 25.}
\label{fig:counts}
\vspace{-0.3cm}
\end{figure*}

We show the plots for the per-image total object count distribution in the \texttt{train} and \texttt{val} splits of our COST dataset in \cref{fig:counts}. We observe that there exists a long tail beyond the object count of 25. Based on this observation, we express the need for a more scaled effort at collecting object-level perception datasets for training MLLMs to make them excel (without extra pre-processing) at counting in cluttered scenes that may contain many more objects.

\end{document}